\documentclass[11pt]{article}
\pdfoutput=1

\usepackage[preprint]{acl}

\usepackage{times}
\usepackage{latexsym}
\usepackage[T1]{fontenc}
\usepackage[utf8]{inputenc}
\usepackage{microtype}
\usepackage{inconsolata}

\usepackage{graphicx}
\graphicspath{{figures/}}
\usepackage{booktabs}
\usepackage{multirow}
\usepackage{makecell}

\usepackage{amsmath}
\usepackage{amsfonts}
\usepackage{amssymb}

\usepackage{enumitem}
\usepackage{xcolor}
\usepackage{subcaption}

\newcommand{\method}{\textsc{DiscoUQ}}
\newcommand{\methodfull}{Disagreement-Structure Confidence for Uncertainty Quantification}
\newcommand{\best}[1]{\textbf{#1}}

\title{\method{}: Structured Disagreement Analysis for\\Uncertainty Quantification in LLM Agent Ensembles}

\author{Bo Jiang \\
  Temple University \\
  \texttt{bo.jiang@temple.edu}}

\begin{document}
\maketitle

\begin{abstract}
Multi-agent LLM systems, where multiple prompted instances of a language model independently answer questions, are increasingly used for complex reasoning tasks.
However, existing methods for quantifying the uncertainty of their collective outputs rely on shallow voting statistics that discard the rich semantic information in agents' reasoning.
We introduce \method{}, a framework that extracts and leverages the \emph{structure} of inter-agent disagreement---both linguistic properties (evidence overlap, argument strength, divergence depth) and embedding geometry (cluster distances, dispersion, cohesion)---to produce well-calibrated confidence estimates.
We propose three methods of increasing complexity: \method{}-LLM (logistic regression on LLM-extracted structure features), \method{}-Embed (logistic regression on embedding geometry), and \method{}-Learn (a neural network combining all features).
Evaluated on four diverse benchmarks (StrategyQA, MMLU, TruthfulQA, ARC-Challenge) with a 5-agent system using Qwen3.5-27B, \method{}-LLM achieves an average AUROC of 0.802, outperforming the best baseline (LLM Aggregator, 0.791) while being substantially better calibrated (ECE 0.036 vs.\ 0.098).
The learned features generalize across benchmarks with near-zero performance degradation and provide the largest improvements where they are most needed: in the ambiguous ``weak disagreement'' tier where simple vote counting fails.
\end{abstract}

\section{Introduction}
\label{sec:intro}

Multi-agent LLM systems, in which several role-specialized instances of a language model independently reason about the same question, have emerged as a powerful paradigm for improving both accuracy and robustness~\cite{du2024improving,liang2024encouraging,li2024more}.
The standard approach to aggregating their outputs is majority voting, a strategy grounded in the Condorcet jury theorem~\cite{condorcet1785essay} and the wisdom-of-crowds literature~\cite{surowiecki2005wisdom}.
When agents disagree, majority voting assigns a confidence proportional to the vote margin---3-to-2 yields 60\%, 4-to-1 yields 80\%.

This approach, however, treats all 3-to-2 disagreements identically, discarding the rich semantic information in agents' reasoning.
Consider two questions where three out of five agents agree: in one case, the minority agents use entirely different evidence and produce weak arguments; in the other, the minority introduces compelling new information that the majority overlooked, and all agents share the same evidence base but diverge at the final conclusion.
These two scenarios carry very different implications for the reliability of the majority vote, yet vote counting assigns them the same confidence.

We introduce \method{} (\methodfull{}), a framework that moves beyond vote counting by analyzing the \emph{structure} of inter-agent disagreement.
Our key insight is that disagreement among LLM agents has a rich internal structure---characterized by what evidence agents share, where in their reasoning they diverge, how strong the minority's arguments are, and how the agents' representations cluster geometrically---that is highly informative of whether the majority vote is correct.

\method{} extracts two complementary families of features from multi-agent outputs:
\textbf{(1) Linguistic structure features}, obtained by prompting the LLM to analyze the disagreement pattern (evidence overlap, minority argument strength, divergence depth, etc.), and
\textbf{(2) Embedding geometry features}, computed from the cosine-distance structure of agent reasoning embeddings (cluster distance, dispersion, cohesion, etc.).
These features are combined through simple classifiers to produce calibrated confidence estimates.

We make the following contributions:
\begin{itemize}[nosep,leftmargin=*]
\item We formalize the notion of \emph{disagreement structure} in multi-agent LLM systems as a set of linguistically and geometrically motivated features.
\item We propose three methods of increasing complexity---\method{}-LLM, \method{}-Embed, and \method{}-Learn---that use these features for uncertainty quantification.
\item We conduct extensive experiments on four benchmarks with nine methods, showing that \method{}-LLM achieves the best average AUROC (0.802) with the best calibration (ECE 0.036), and that the features generalize across benchmarks.
\end{itemize}

\section{Related Work}
\label{sec:related}

\paragraph{Uncertainty quantification in LLMs.}
Estimating the reliability of LLM outputs is a growing area of research.
Verbalized confidence~\cite{kadavath2022language,tian2023just,xiong2024can} asks models to self-report their certainty, but LLMs are often poorly calibrated and tend toward overconfidence.
Token-level methods use output probabilities~\cite{jiang2021how}, but these are unavailable through many API interfaces.
Semantic uncertainty~\cite{kuhn2023semantic} clusters sampled outputs by meaning and measures the entropy, while SelfCheckGPT~\cite{manakul2023selfcheckgpt} detects hallucinations by checking consistency across samples.
Lin et al.~\shortcite{lin2024generating} and Ye et al.~\shortcite{ye2024benchmarking} provide broader surveys of black-box UQ methods.
Our work differs by leveraging the \emph{structure} of disagreement rather than just its presence.

\paragraph{Multi-agent debate and aggregation.}
Du et al.~\shortcite{du2024improving} show that multi-agent debate improves factuality, and Liang et al.~\shortcite{liang2024encouraging} demonstrate that diverse agent roles encourage broader reasoning.
Cemri et al.~\shortcite{cemri2025mast} provide a taxonomy of multi-agent system failure modes (MAST), identifying communication breakdowns and cascading errors; our work is complementary, quantifying \emph{when} such failures are likely rather than categorizing them post-hoc.
Kim et al.~\shortcite{kim2025scaling} study how multi-agent architectures scale across configurations but do not address uncertainty estimation.
These works focus on improving the \emph{accuracy} or understanding the \emph{failures} of multi-agent systems; our focus is on quantifying the \emph{uncertainty} of the consensus.

\paragraph{Ensemble methods and calibration.}
Deep ensembles~\cite{lakshminarayanan2017simple} use prediction disagreement for uncertainty estimation, drawing on classical ensemble theory~\cite{dietterich2000ensemble}.
Calibration of neural networks has been widely studied~\cite{guo2017calibration,naeini2015obtaining}.
Self-consistency~\cite{wang2023selfconsistency} applies ensemble-like reasoning to chain-of-thought prompting.
Our work extends the ensemble paradigm by moving beyond vote counting to analyze the \emph{semantic content} of disagreement.

\paragraph{Selective prediction.}
Selective prediction~\cite{geifman2017selective,kamath2020selective} allows a model to abstain when uncertain.
Our confidence estimates directly enable selective prediction, and we evaluate this through coverage metrics.

\section{Method}
\label{sec:method}

\subsection{Multi-Agent System}
\label{sec:multi-agent}

Given a question $q$ with ground truth $y^*$, we deploy $K{=}5$ role-specialized agents $\{a_1, \ldots, a_K\}$, each with a distinct system prompt encouraging a different reasoning style (Analytical Reasoner, Devil's Advocate, Knowledge-Focused, Intuitive Responder, Systematic Verifier).
All agents independently produce a reasoning chain $r_k$ and a final answer $\hat{y}_k$.
The majority answer $\hat{y} = \text{mode}(\hat{y}_1, \ldots, \hat{y}_K)$ and the vote confidence $c_\text{vote} = |\{k : \hat{y}_k = \hat{y}\}| / K$ form the baseline.

\subsection{Disagreement Structure Features}
\label{sec:features}

We extract two complementary feature families from the agents' outputs.

\subsubsection{Linguistic Structure Features}
\label{sec:structure}

For non-unanimous questions ($c_\text{vote} < 1$), we prompt the same LLM (with temperature 0) to analyze the disagreement among the $K$ agents' reasoning texts and produce six scores:

\begin{itemize}[nosep,leftmargin=*]
\item \textbf{Evidence overlap} ($e_\text{overlap} \in [0,1]$): degree to which majority and minority agents cite the same facts or evidence.
\item \textbf{Minority new information} ($e_\text{new} \in [0,1]$): extent of genuinely new arguments introduced by the minority.
\item \textbf{Minority argument strength} ($e_\text{str} \in [0,1]$): logical soundness of the minority's reasoning, independent of correctness.
\item \textbf{Majority confidence language} ($e_\text{conf} \in [0,1]$): certainty expressed in the majority's language (hedging vs.\ assertive).
\item \textbf{Reasoning complexity} ($e_\text{cplx} \in [0,1]$): complexity of the overall reasoning across agents.
\item \textbf{Divergence depth} ($d \in \{\text{early}, \text{middle}, \text{late}\}$): the stage at which agents' reasoning diverges---from the initial framing, during intermediate steps, or only at the final conclusion.
\end{itemize}

For unanimous votes, we assign default values ($e_\text{overlap}{=}1$, $e_\text{str}{=}0$, $d{=}\text{none}$).
The categorical divergence depth is one-hot encoded, yielding 9 total LLM structure features (including $c_\text{vote}$).

\subsubsection{Embedding Geometry Features}
\label{sec:geometry}

We embed each agent's reasoning text $r_k$ using a sentence transformer (BGE-large, 1024-dim; \citealt{xiao2024bge}) and compute seven geometric features from the cosine-distance structure of the resulting embedding vectors:

\begin{itemize}[nosep,leftmargin=*]
\item \textbf{Overall dispersion}: mean pairwise cosine distance across all $K$ agents.
\item \textbf{Majority cohesion}: mean pairwise distance within the majority group.
\item \textbf{Cluster distance}: distance between majority and minority centroids.
\item \textbf{Minority outlier degree}: mean distance from minority agents to majority centroid.
\item \textbf{Majority centrality}: distance from majority centroid to the global centroid.
\item \textbf{Minority cohesion}: mean pairwise distance within the minority group.
\item \textbf{PCA variance ratio}: first principal component's explained variance ratio.
\end{itemize}

Together with $c_\text{vote}$, this yields 8 embedding features.

\subsection{Uncertainty Quantification Methods}
\label{sec:uq-methods}

We propose three methods of increasing complexity, all predicting $P(\hat{y} = y^* \mid q)$:

\paragraph{M1: \method{}-LLM.}
Logistic regression on the 9 LLM structure features (vote confidence + 5 continuous structure scores + 3 one-hot divergence depth indicators).
This method is highly interpretable: each coefficient directly reveals which disagreement patterns correlate with correctness.

\paragraph{M2: \method{}-Embed.}
Logistic regression on the 8 embedding geometry features (vote confidence + 7 geometric measures).
This method requires no additional LLM calls beyond the initial agent queries, since embeddings are extracted via a separate encoder.

\paragraph{M3: \method{}-Learn.}
A two-layer MLP (32 hidden units, ReLU, dropout 0.3) on all 17 features (vote confidence, mean verbalized confidence, 5 structure scores, 3 divergence depth indicators, 7 geometry features).
Trained with binary cross-entropy loss, Adam optimizer (lr=$5{\times}10^{-4}$, weight decay=$10^{-2}$), and early stopping (patience 15, max 100 epochs).

All methods are evaluated via 5-fold stratified cross-validation.

\section{Experimental Setup}
\label{sec:experiments}

\subsection{Benchmarks}

We evaluate on four diverse question-answering benchmarks:

\begin{itemize}[nosep,leftmargin=*]
\item \textbf{StrategyQA}~\cite{geva2021strategyqa}: 687 yes/no questions requiring implicit multi-step reasoning.
\item \textbf{MMLU}~\cite{hendrycks2021measuring}: 746 multiple-choice questions from three subjects (logical fallacies, philosophy, professional medicine).
\item \textbf{TruthfulQA}~\cite{lin2022truthfulqa}: 817 multiple-choice questions designed to test resistance to common misconceptions.
\item \textbf{ARC-Challenge}~\cite{clark2018think}: 500 multiple-choice science questions from the Challenge set.
\end{itemize}

In total, we evaluate on 2,750 questions spanning yes/no and multiple-choice formats, covering knowledge retrieval, multi-step reasoning, anti-hallucination, and science reasoning.

\subsection{Models}

Our primary agent model is Qwen3.5-27B~\cite{yang2025qwen3}, served via vLLM with temperature 0.7 and max tokens 800 for agent responses.
For embedding extraction, we use BGE-large-en-v1.5~\cite{xiao2024bge} (1024 dimensions).
For the heterogeneous agent experiment (\S\ref{sec:heterogeneous}), we additionally use Llama-3.1-8B-Instruct~\cite{grattafiori2024llama3}.

\subsection{Baselines}

We compare against six baselines:

\begin{itemize}[nosep,leftmargin=*]
\item \textbf{B1: Vote Count.} $c = |\text{majority}| / K$.
\item \textbf{B2: Vote Entropy.} $c = 1 - H(\hat{\mathbf{y}}) / \log_2 |\mathcal{Y}|$, where $H$ is Shannon entropy.
\item \textbf{B3: Verbalized Confidence.} Mean self-reported confidence of majority agents (0--100 scale, via follow-up prompts).
\item \textbf{B4: Self-Consistency Entropy.} Answer entropy with normalization~\cite{wang2023selfconsistency}.
\item \textbf{B5: Embed Centroid.} $c = 1 - d_\text{cos}(\text{majority centroid}, \text{global centroid})$.
\item \textbf{B6: LLM Aggregator.} A single LLM call that reads all five agent responses and outputs its own answer and confidence in JSON format.
\end{itemize}

B6 is a strong baseline that has access to the same information as our methods but processes it through a single forward pass rather than structured feature extraction.
Note that B6 selects its own answer (which may differ from the majority vote), so its correctness is judged against B6's answer.

\subsection{Evaluation Metrics}

We evaluate using seven complementary metrics:
\textbf{AUROC} (primary metric, discrimination ability),
\textbf{ECE} (expected calibration error, 10 bins; \citealt{naeini2015obtaining}),
\textbf{Brier score}~\cite{brier1950verification},
\textbf{AUPRC} (area under precision-recall curve),
\textbf{Coverage@90/95} (fraction of data retained at $\geq$90\%/95\% accuracy),
and \textbf{AUACC} (area under accuracy-coverage curve; \citealt{geifman2017selective}).
Statistical significance is assessed via paired bootstrap (1,000 resamples, $\alpha{=}0.05$).

\section{Results}
\label{sec:results}

\subsection{Main Results}

Table~\ref{tab:main} presents the AUROC and ECE for all nine methods across four benchmarks.

\begin{table*}[t]
\centering
\small
\begin{tabular}{l cc cc cc cc cc}
\toprule
& \multicolumn{2}{c}{\textbf{StrategyQA}} & \multicolumn{2}{c}{\textbf{MMLU}} & \multicolumn{2}{c}{\textbf{TruthfulQA}} & \multicolumn{2}{c}{\textbf{ARC-Chal.}} & \multicolumn{2}{c}{\textbf{Average}} \\
\cmidrule(lr){2-3} \cmidrule(lr){4-5} \cmidrule(lr){6-7} \cmidrule(lr){8-9} \cmidrule(lr){10-11}
\textbf{Method} & AUC & ECE & AUC & ECE & AUC & ECE & AUC & ECE & AUC & ECE \\
\midrule
B1: Vote Count & .647 & .095 & .841 & .114 & .798 & .037 & .796 & .089 & .770 & .084 \\
B2: Vote Entropy & .647 & .338 & .837 & .205 & .798 & .096 & .794 & .185 & .769 & .206 \\
B3: Verb.\ Conf. & .681 & .205 & .720 & .143 & .733 & .122 & .760 & .042 & .724 & .128 \\
B4: Self-Consist. & .647 & .338 & .837 & .205 & .798 & .096 & .794 & .185 & .769 & .206 \\
B5: Embed Centr. & .657 & .244 & .845 & .186 & .807 & .155 & .778 & .056 & .772 & .160 \\
B6: LLM Agg. & .670 & .186 & .822 & .059 & .805 & .102 & \best{.868} & .044 & .791 & .098 \\
\midrule
M1: \method{}-LLM & \best{.706} & \best{.008} & \best{.870} & .049 & .867 & .062 & .766 & \best{.024} & \best{.802} & \best{.036} \\
M2: \method{}-Embed & .676 & .026 & .840 & .056 & .798 & .039 & .779 & .039 & .773 & .040 \\
M3: \method{}-Learn & .689 & .038 & .867 & .048 & \best{.871} & .074 & .731 & .009 & .790 & .042 \\
\bottomrule
\end{tabular}
\caption{AUROC (higher is better) and ECE (lower is better) across four benchmarks. \textbf{Bold}: best per column. \method{}-LLM achieves the highest average AUROC (0.802) and the lowest average ECE (0.036). Bootstrap 95\% CIs and significance tests are in Appendix~\ref{sec:full-metrics}.}
\label{tab:main}
\end{table*}

\method{}-LLM (M1) achieves the highest average AUROC (0.802), significantly outperforming all baselines including B6 LLM Aggregator (0.791, $p{<}0.05$ on StrategyQA, MMLU, and TruthfulQA).
\method{}-Learn (M3) is the second-best method overall (0.790), while \method{}-Embed (M2) performs on par with the best non-LLM baseline B5 (0.773 vs.\ 0.772) but with dramatically better calibration (ECE 0.040 vs.\ 0.160).

All three \method{} methods achieve substantially better calibration than baselines.
The average ECE drops from 0.084 (B1) and 0.098 (B6) to 0.036 (M1), representing a 57\% and 63\% reduction, respectively.
This suggests that structured disagreement features naturally produce well-calibrated probabilities through logistic regression's sigmoid function.

B6 LLM Aggregator is noteworthy as the strongest individual-benchmark performer on ARC-Challenge (0.868), likely because it can leverage the full reasoning of all agents through a single forward pass.
However, it is poorly calibrated on most benchmarks, confirming that raw LLM confidence scores require post-hoc adjustment.

\subsection{Performance by Disagreement Tier}
\label{sec:tier-analysis}

\begin{figure}[t]
\centering
\includegraphics[width=\columnwidth]{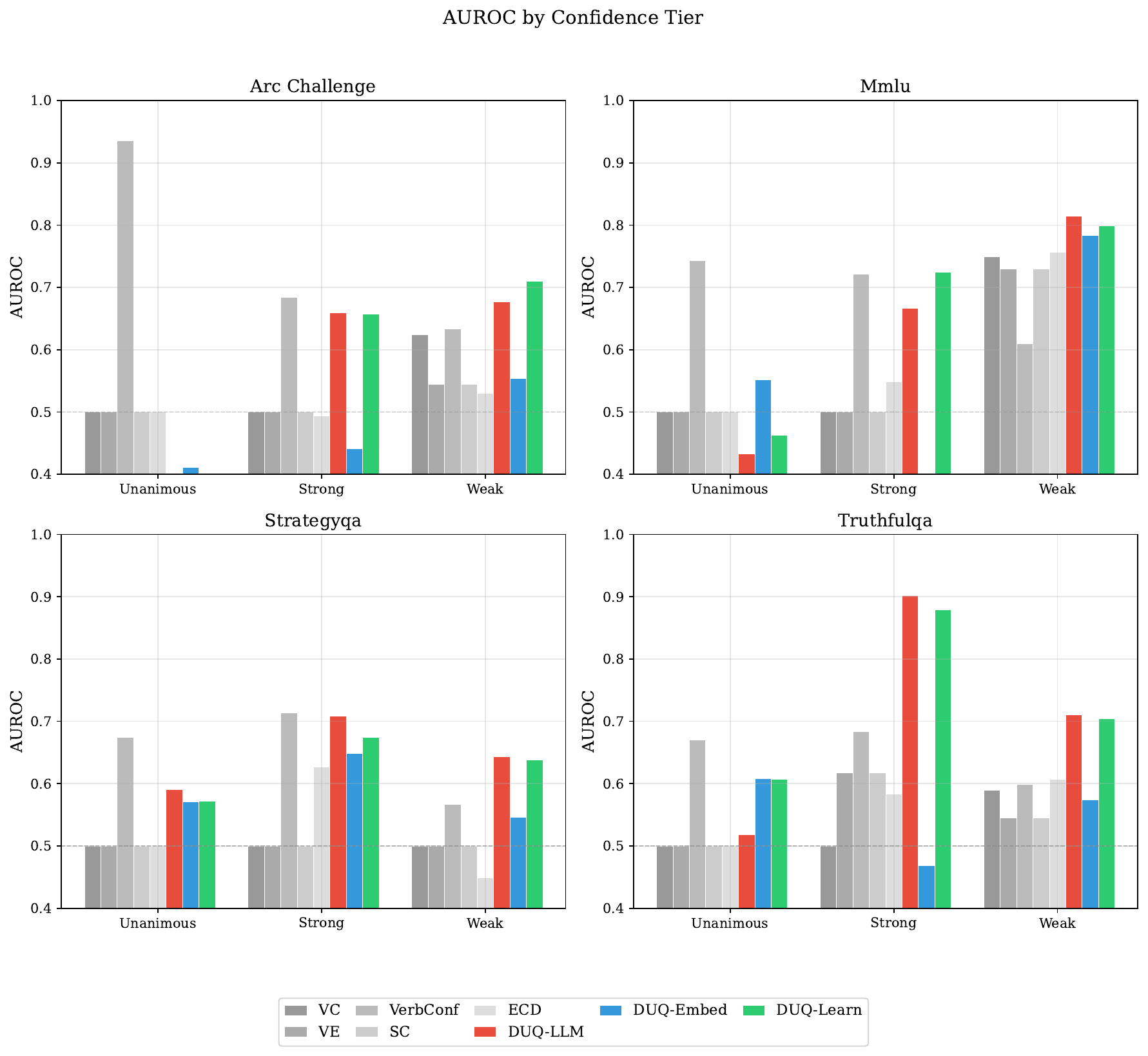}
\caption{AUROC by disagreement tier across benchmarks. \method{} methods show the largest improvements in the strong and weak tiers where vote counting provides the least information.}
\label{fig:auroc-tier}
\end{figure}

We stratify questions into three tiers based on vote confidence: \emph{unanimous} ($c_\text{vote}{=}1.0$), \emph{strong} ($c_\text{vote}{\geq}0.8$), and \emph{weak} ($c_\text{vote}{<}0.8$).
Figure~\ref{fig:auroc-tier} shows AUROC by tier and benchmark.

In the weak tier, where agents are most divided and vote counting is least informative, \method{} methods provide substantial gains.
On MMLU's weak tier (252 questions), M1 achieves AUROC 0.814 vs.\ B1's 0.749, a 6.5 percentage point improvement.
On TruthfulQA's strong tier (243 questions), M1 reaches AUROC 0.902, demonstrating that linguistic structure features capture fine-grained disagreement signals even when agents narrowly agree.

However, B6 LLM Aggregator is the strongest method in the weak tier overall (average weak-tier AUROC 0.736 vs.\ M1's 0.711), because it can read the full reasoning text of all five agents and re-derive an answer---an informational advantage when disagreement is extreme.
Notably, M1 approaches this upper bound using only six scalar features extracted from the same text, suggesting that the structured decomposition captures most of the signal available in the raw reasoning.

In the unanimous tier, where all agents agree, vote counting assigns $c{=}1.0$ and thus AUROC$=$0.5.
Even here, M1 can discriminate correct from incorrect unanimous answers (e.g., AUROC 0.59 on StrategyQA) by leveraging verbalized confidence and structure features from a prior analysis pass.

\subsection{Calibration Analysis}

\begin{figure}[t]
\centering
\includegraphics[width=\columnwidth]{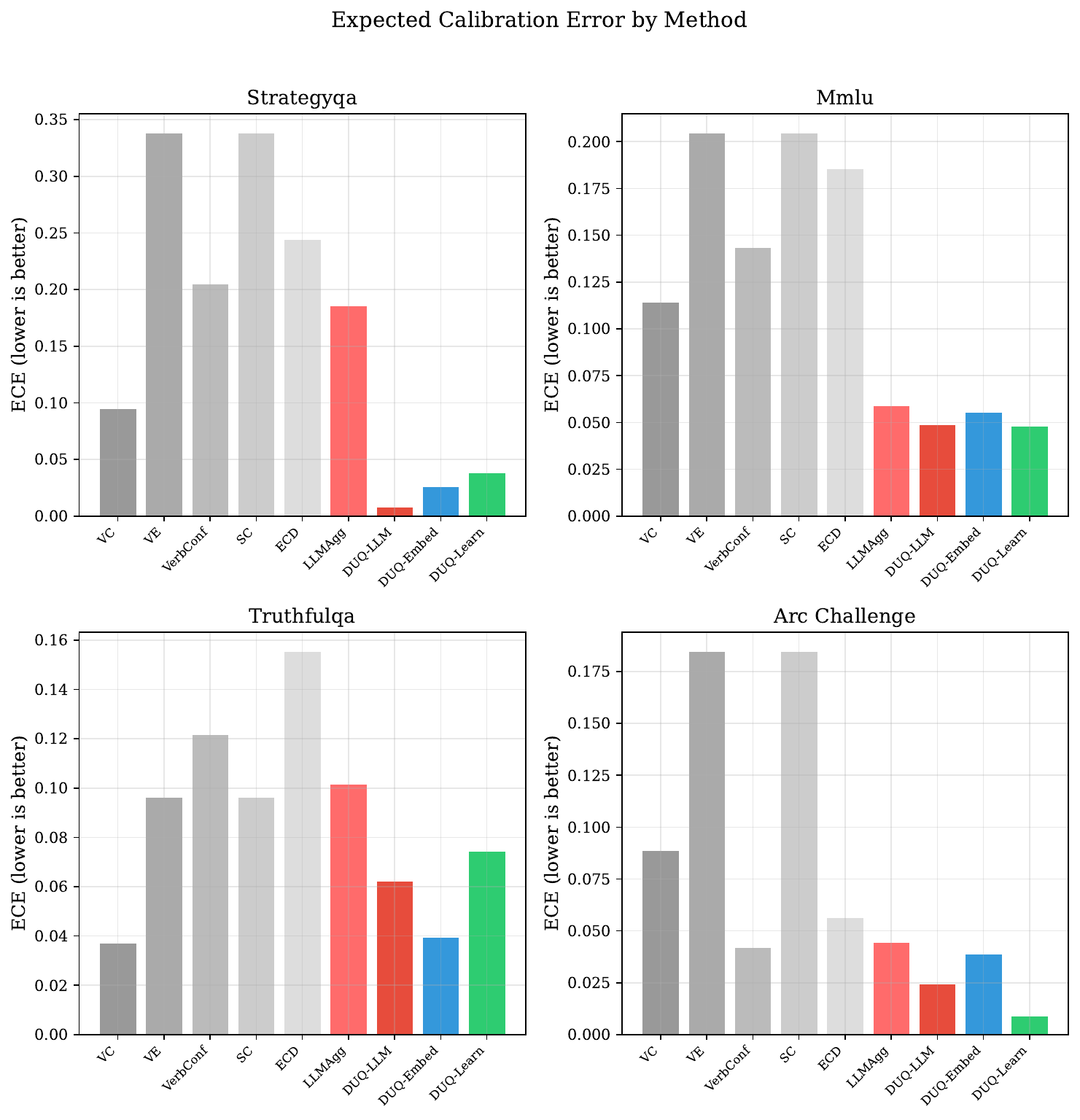}
\caption{Expected calibration error (ECE) across benchmarks. Lower is better. All \method{} methods achieve substantially lower ECE than baselines.}
\label{fig:calibration}
\end{figure}

Figure~\ref{fig:calibration} shows the ECE comparison across benchmarks.

The \method{} methods achieve ECE $\leq$ 0.074 across all benchmarks, while baselines range from 0.037 to 0.338.
The improvement is most dramatic on StrategyQA (M1: ECE 0.008 vs.\ B1: 0.095) and ARC-Challenge (M3: ECE 0.009 vs.\ B1: 0.089).
This reflects the fact that logistic regression naturally produces calibrated probabilities when the features are informative, whereas heuristic baselines have no mechanism for calibration.

\subsection{Selective Prediction}

\begin{figure}[t]
\centering
\includegraphics[width=\columnwidth]{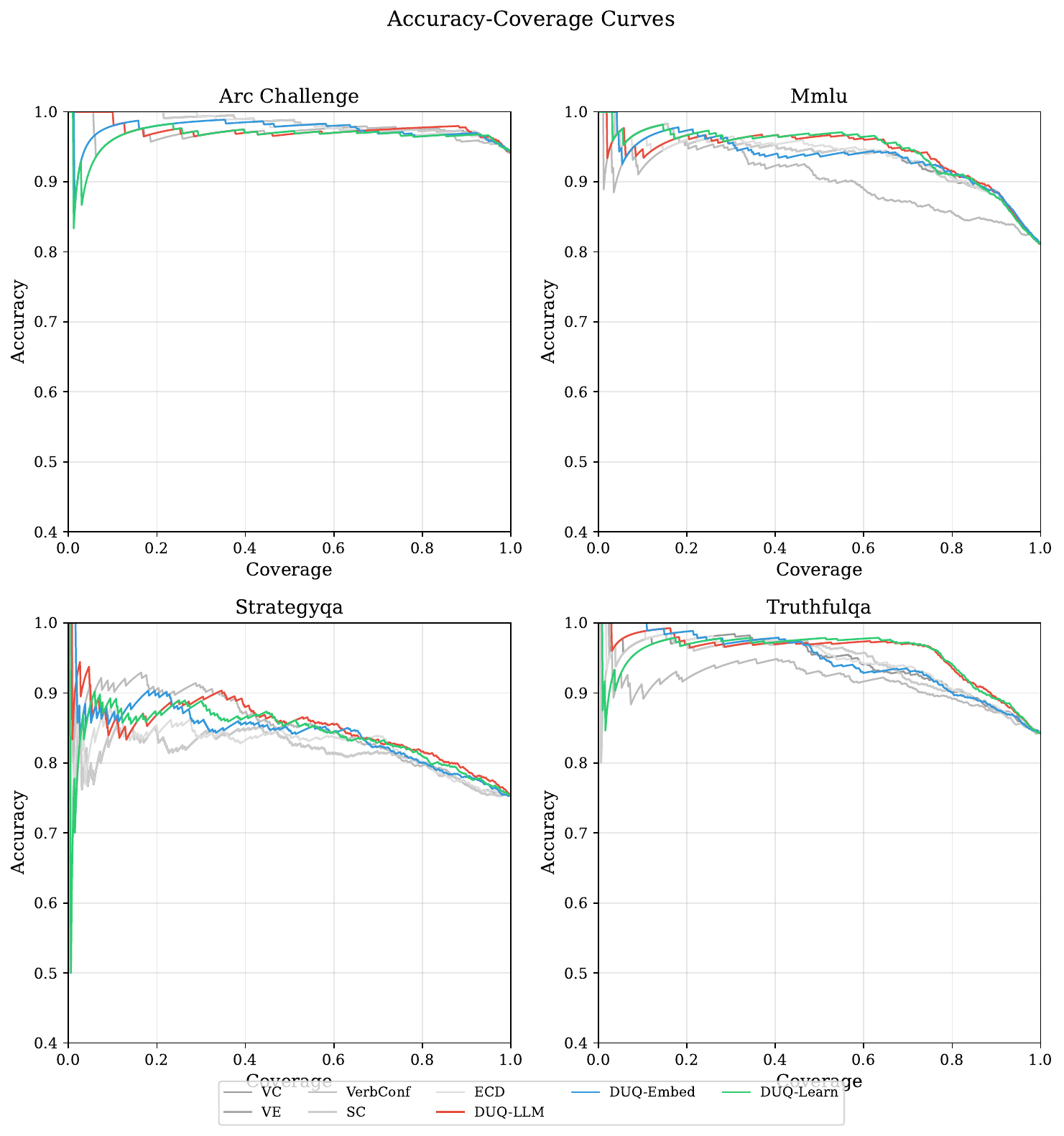}
\caption{Accuracy-coverage curves across benchmarks. \method{} methods maintain higher accuracy at all coverage levels, particularly on StrategyQA and TruthfulQA.}
\label{fig:acc-cov}
\end{figure}

Figure~\ref{fig:acc-cov} shows accuracy-coverage curves, which trace the tradeoff between abstaining on uncertain questions and maintaining high accuracy on the remaining ones.

On MMLU, M1 achieves Coverage@95 of 0.656 (i.e., 65.6\% of questions can be answered at $\geq$95\% accuracy), compared to 0.404 for B1 Vote Count.
On TruthfulQA, M1 achieves Coverage@95 of 0.775 vs.\ B1's 0.572.
These improvements demonstrate the practical value of \method{} for selective prediction: systems can serve substantially more questions while maintaining their accuracy guarantee.

\section{Analysis}
\label{sec:analysis}

\subsection{Cross-Benchmark Generalization}
\label{sec:cross-bm}

A key question is whether the disagreement patterns learned on one set of questions transfer to others.
We evaluate this with leave-one-benchmark-out training: train on three benchmarks (pooling all questions) and test on the fourth.

\begin{table}[t]
\centering
\small
\begin{tabular}{l cccc c}
\toprule
& \textbf{Str.} & \textbf{MMLU} & \textbf{TQA} & \textbf{ARC} & \textbf{Avg.} \\
\midrule
\multicolumn{6}{l}{\emph{M1: \method{}-LLM}} \\
\quad Cross-BM & .679 & .870 & .856 & .801 & .802 \\
\quad Same-BM & .706 & .870 & .867 & .766 & .802 \\
\quad $\Delta$ & -.027 & +.001 & -.011 & +.035 & -.001 \\
\midrule
\multicolumn{6}{l}{\emph{M3: \method{}-Learn}} \\
\quad Cross-BM & .696 & .892 & .870 & .839 & .824 \\
\quad Same-BM & .687 & .868 & .876 & .695 & .781 \\
\quad $\Delta$ & +.009 & +.025 & -.006 & +.144 & +.043 \\
\bottomrule
\end{tabular}
\caption{Cross-benchmark generalization (AUROC). Cross-BM: trained on 3 benchmarks, tested on the held-out one. Same-BM: within-benchmark 5-fold CV. $\Delta = \text{Cross} - \text{Same}$.}
\label{tab:cross-bm}
\end{table}

Table~\ref{tab:cross-bm} shows that M1's cross-benchmark AUROC (0.802) is virtually identical to its within-benchmark performance (0.802, $\Delta{=}{-}0.001$), indicating that the structure features capture domain-general patterns of disagreement informativeness.
M3 actually \emph{improves} under cross-benchmark training ($\Delta{=}{+}0.043$), benefiting from the larger and more diverse training set (2,000+ vs.\ $\sim$500 questions per fold).
All cross-benchmark AUROCs exceed 0.5, confirming that the features are universally informative.

\subsection{Feature Ablation}
\label{sec:ablation}

We conduct ablation studies on StrategyQA (Table~\ref{tab:ablation}).

\begin{table}[t]
\centering
\small
\begin{tabular}{l c}
\toprule
\textbf{Configuration} & \textbf{AUROC} \\
\midrule
\multicolumn{2}{l}{\emph{M1: LLM Features}} \\
\quad Full (9 features) & .706 \\
\quad $-$ majority confidence lang. & .683 \scriptsize{($-$.023)} \\
\quad $-$ reasoning complexity & .699 \scriptsize{($-$.007)} \\
\quad $-$ vote confidence & .699 \scriptsize{($-$.007)} \\
\quad $-$ divergence depth (late) & .702 \scriptsize{($-$.004)} \\
\quad Vote conf.\ only & .637 \scriptsize{($-$.069)} \\
\midrule
\multicolumn{2}{l}{\emph{M2: Embedding Features}} \\
\quad Full (8 features) & .676 \\
\quad $-$ PCA variance ratio & .658 \scriptsize{($-$.018)} \\
\quad $-$ minority outlier degree & .674 \scriptsize{($-$.002)} \\
\quad Vote conf.\ only & .637 \scriptsize{($-$.039)} \\
\midrule
\multicolumn{2}{l}{\emph{Agent count (M1)}} \\
\quad 3 agents & .569 \\
\quad 4 agents & .617 \\
\quad 5 agents & .637 \\
\bottomrule
\end{tabular}
\caption{Ablation on StrategyQA. For M1, majority confidence language is the most important individual feature. For M2, PCA variance ratio contributes the most beyond vote confidence.}
\label{tab:ablation}
\end{table}

For M1, the single most important feature beyond vote confidence is \emph{majority confidence language} ($\Delta{=}{-}0.023$), which measures how certain the majority sounds in their language.
This aligns with the intuition that hedging language from confident-seeming majorities signals potential unreliability.
The full set of structure features provides a 6.9 percentage point improvement over vote confidence alone (0.706 vs.\ 0.637), confirming that disagreement structure is highly informative.

For M2, PCA variance ratio ($\Delta{=}{-}0.018$) is the most informative geometry feature, suggesting that the degree to which agent embeddings align along a single dimension captures meaningful disagreement structure.

Agent count ablation shows a clear relationship: more agents provide richer disagreement signals (3 agents: 0.569, 5 agents: 0.637).

\paragraph{Classifier choice.}
We also compare classifiers on M1's features: Logistic Regression (0.706), Random Forest (0.730), SVM-RBF (0.698).
While Random Forest slightly outperforms LR, we use LR throughout for its interpretability and calibration properties.

\subsection{Cost-Performance Analysis}
\label{sec:cost}

\begin{figure}[t]
\centering
\includegraphics[width=\columnwidth]{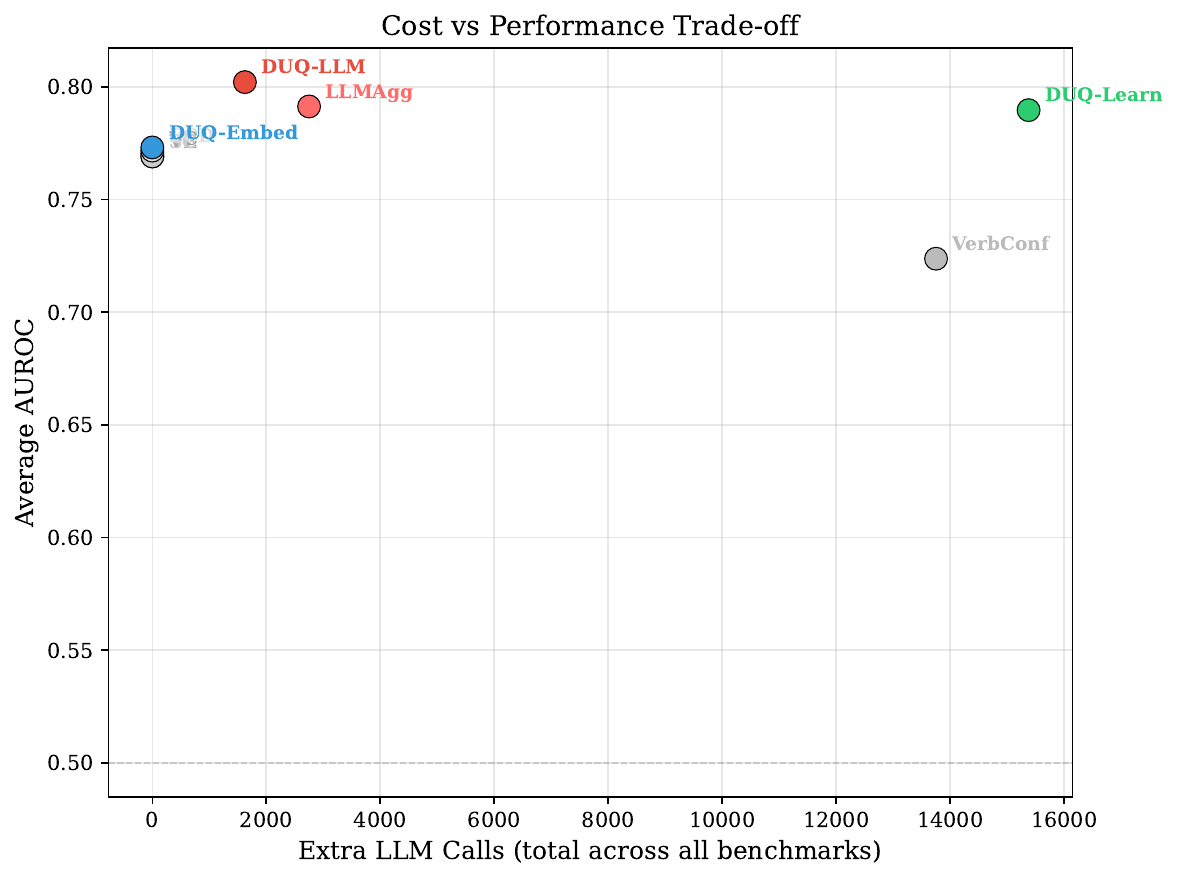}
\caption{Cost (extra LLM calls) vs.\ performance (average AUROC) for all methods. M1 offers the best cost-performance tradeoff.}
\label{fig:cost}
\end{figure}

Figure~\ref{fig:cost} and Table~\ref{tab:cost} show the computational overhead of each method.

\begin{table}[t]
\centering
\small
\begin{tabular}{l r r c}
\toprule
\textbf{Method} & \textbf{Extra calls} & \textbf{Tok/q} & \textbf{AUROC} \\
\midrule
B1: Vote Count & 0 & 0 & .770 \\
M2: \method{}-Embed & 0 & 0 & .773 \\
M1: \method{}-LLM & 1,624 & 2,799 & \best{.802} \\
B6: LLM Agg. & 2,750 & 1,387 & .791 \\
B3: Verb.\ Conf. & 13,750 & 623 & .724 \\
M3: \method{}-Learn & 15,374 & 3,422 & .790 \\
\bottomrule
\end{tabular}
\caption{Computational cost vs.\ AUROC. M1 achieves the best AUROC with only 1,624 extra calls (structure analysis for non-unanimous questions only). M2 requires zero extra LLM calls.}
\label{tab:cost}
\end{table}

M1 is remarkably cost-efficient: it requires only 1,624 extra LLM calls (one per non-unanimous question) to achieve the best AUROC of 0.802.
B6 requires 2,750 calls (one per question regardless) yet achieves a lower AUROC (0.791).
M2 requires \emph{zero} extra LLM calls (only embedding model inference) and still outperforms all baselines except B6.

M3's higher cost (15,374 calls) stems from requiring both structure analysis and verbalized confidence.
Despite this, its AUROC (0.790) does not exceed M1's, suggesting that the additional features and model complexity are not worth the cost in this regime.

\subsection{Heterogeneous Agent Teams}
\label{sec:heterogeneous}

We investigate whether mixing different LLMs (Qwen3.5-27B and Llama-3.1-8B) as agents affects disagreement structure.
In the heterogeneous configuration, agents 0, 2, 4 use Qwen and agents 1, 3 use Llama.

\begin{table}[t]
\centering
\small
\begin{tabular}{l cc cc}
\toprule
& \multicolumn{2}{c}{\textbf{StrategyQA}} & \multicolumn{2}{c}{\textbf{TruthfulQA}} \\
\cmidrule(lr){2-3} \cmidrule(lr){4-5}
& Homo & Hetero & Homo & Hetero \\
\midrule
Disagree. rate & 53.0\% & 45.6\% & 54.3\% & 49.2\% \\
\midrule
B1: Vote Count & .647 & .675 & .798 & .881 \\
M1: \method{}-LLM & .706 & .687 & .867 & .904 \\
\bottomrule
\end{tabular}
\caption{Homogeneous vs.\ heterogeneous agent teams. Heterogeneous teams show \emph{lower} disagreement rates but higher B1 performance, suggesting that cross-model disagreement is more informative.}
\label{tab:hetero}
\end{table}

Counter-intuitively, heterogeneous teams show \emph{lower} disagreement rates (StrategyQA: 45.6\% vs.\ 53.0\%; TruthfulQA: 49.2\% vs.\ 54.3\%).
This likely occurs because Llama-3.1-8B is weaker and tends to agree with the majority more often.
However, when heterogeneous agents \emph{do} disagree, the disagreement is more informative: B1 improves substantially on TruthfulQA (0.798 $\rightarrow$ 0.881).
M1 also improves on TruthfulQA (0.867 $\rightarrow$ 0.904), suggesting that cross-model disagreement carries additional signal.

\section{Discussion}
\label{sec:discussion}

\paragraph{Why does structure help?}
The core finding is that disagreement has meaningful internal structure that vote counting discards.
When the minority introduces genuinely new evidence ($e_\text{new}$ high) and makes a strong argument ($e_\text{str}$ high), the majority vote is less reliable---even if the vote margin is the same.
Conversely, when agents diverge only at the final conclusion ($d{=}\text{late}$) but share evidence ($e_\text{overlap}$ high), the majority is typically correct.
These patterns are captured by simple logistic regression, explaining M1's strong performance without complex modeling.

\paragraph{Calibration advantage.}
The dramatic calibration improvement (ECE 0.036 vs.\ 0.084) arises because logistic regression's sigmoid function naturally maps informative features to well-calibrated probabilities.
In contrast, vote counting produces only $K{+}1$ discrete confidence values, making calibration inherently coarse.

\paragraph{When does \method{} underperform?}
On ARC-Challenge, where the base accuracy is very high (96\%), M1 and M3 underperform B1 and B6 in AUROC.
With so few errors, there is limited room for disagreement-based features to discriminate, and B6's access to full agent reasoning gives it an advantage on this nearly-saturated benchmark.
M3's performance drop on ARC (0.731 vs.\ B1's 0.796) is particularly notable: with only ${\sim}20$ incorrect unanimous questions, the MLP overfits to noise in the minority class, producing miscalibrated confidence on the test fold.
M1's simpler logistic regression is more robust to this low-error regime (0.766), though it still trails the heuristic baselines.

\paragraph{Practical deployment.}
M1 offers the best cost-performance tradeoff for production systems: it requires only one extra LLM call per ambiguous question (1,624 calls for 2,750 questions) while achieving the best AUROC and calibration.
For cost-sensitive applications, M2 achieves strong calibration (ECE 0.040) with zero additional LLM calls.

\section{Conclusion}

We introduced \method{}, a framework for uncertainty quantification in multi-agent LLM systems that analyzes the structure of inter-agent disagreement.
By extracting linguistic structure features (evidence overlap, argument strength, divergence depth) and embedding geometry features (cluster distances, dispersion), \method{} produces confidence estimates that are both more discriminative and better calibrated than existing approaches.

Our experiments on four benchmarks demonstrate that \method{}-LLM achieves an average AUROC of 0.802 with ECE of 0.036, outperforming the strongest baseline (LLM Aggregator: AUROC 0.791, ECE 0.098) while requiring fewer LLM calls.
The features generalize across benchmarks with negligible performance loss, and provide the largest improvements where they are most needed---in ambiguous cases where agents are divided.

This work opens several directions for future research: extending structure analysis to open-ended generation tasks, exploring how disagreement patterns vary across model families, and developing online calibration methods that adapt structure weights over time.

\section*{Limitations}

Our study has several limitations.
\textbf{(1)} We evaluate only on closed-form QA tasks (yes/no and multiple-choice); extending to open-ended generation would require different evaluation criteria.
\textbf{(2)} Our experiments use a single primary model (Qwen3.5-27B); the degree to which disagreement structure generalizes across model families is only partially explored through the heterogeneous experiment.
\textbf{(3)} The LLM-extracted structure features depend on the quality of the analyzing model's introspection, which may degrade for more complex reasoning chains.
\textbf{(4)} Our methods require a labeled dataset for training the classifier, though the cross-benchmark results suggest that labels from one domain can be effective for another.
\textbf{(5)} We use 5 agents; scaling to larger agent pools may introduce different disagreement dynamics.

\bibliography{custom}

\appendix

\section{Full Metrics Table}
\label{sec:full-metrics}

Table~\ref{tab:full-metrics} presents the complete set of seven metrics for all nine methods across four benchmarks, with bootstrap 95\% confidence intervals.

\begin{table*}[t]
\centering
\scriptsize
\setlength{\tabcolsep}{3pt}
\begin{tabular}{ll ccccccc}
\toprule
\textbf{Bench.} & \textbf{Method} & \textbf{AUROC} & \textbf{ECE} & \textbf{Brier} & \textbf{AUPRC} & \textbf{Cov@90} & \textbf{Cov@95} & \textbf{AUACC} \\
\midrule
\multirow{9}{*}{\rotatebox{90}{StrategyQA}}
& B1: Vote Count & .647 {\tiny[.603,.693]} & .095 & .187 & .818 & .016 & .001 & .813 \\
& B2: Vote Entropy & .647 {\tiny[.603,.693]} & .338 & .319 & .818 & .016 & .001 & .813 \\
& B3: Verb.\ Conf. & .681 {\tiny[.638,.724]} & .205 & .224 & .852 & .339 & .009 & .849 \\
& B4: Self-Consist. & .647 {\tiny[.603,.693]} & .338 & .319 & .818 & .016 & .001 & .813 \\
& B5: Embed Centr. & .657 {\tiny[.610,.704]} & .244 & .244 & .830 & .003 & .003 & .823 \\
& B6: LLM Agg. & .670 {\tiny[.622,.718]} & .186 & .194 & .852 & .000 & .000 & .836 \\
& M1: DUQ-LLM & \best{.706} {\tiny[.661,.751]} & \best{.008} & \best{.164} & \best{.856} & \best{.352} & .007 & \best{.848} \\
& M2: DUQ-Embed & .676 {\tiny[.629,.723]} & .026 & .173 & .848 & .221 & \best{.016} & .840 \\
& M3: DUQ-Learn & .689 {\tiny[.643,.735]} & .038 & .171 & .846 & .058 & .003 & .838 \\
\midrule
\multirow{9}{*}{\rotatebox{90}{MMLU}}
& B1: Vote Count & .841 {\tiny[.800,.879]} & .114 & .112 & .935 & .834 & .404 & .934 \\
& B2: Vote Entropy & .837 {\tiny[.795,.876]} & .205 & .153 & .937 & .824 & .404 & .934 \\
& B3: Verb.\ Conf. & .720 {\tiny[.674,.760]} & .143 & .162 & .899 & .579 & .283 & .898 \\
& B4: Self-Consist. & .837 {\tiny[.795,.876]} & .205 & .153 & .937 & .824 & .404 & .934 \\
& B5: Embed Centr. & .845 {\tiny[.805,.880]} & .186 & .186 & .945 & .835 & .524 & .934 \\
& B6: LLM Agg. & .822 {\tiny[.763,.874]} & .059 & .071 & .970 & \best{1.00} & .912 & \best{.966} \\
& M1: DUQ-LLM & \best{.870} {\tiny[.830,.904]} & .049 & \best{.096} & .950 & .854 & \best{.656} & .939 \\
& M2: DUQ-Embed & .840 {\tiny[.795,.880]} & .056 & .104 & .944 & .845 & .308 & .932 \\
& M3: DUQ-Learn & .867 {\tiny[.831,.899]} & \best{.048} & .103 & \best{.951} & .851 & .653 & .938 \\
\midrule
\multirow{9}{*}{\rotatebox{90}{TruthfulQA}}
& B1: Vote Count & .798 {\tiny[.762,.835]} & \best{.037} & .111 & .938 & .823 & .572 & .942 \\
& B2: Vote Entropy & .798 {\tiny[.759,.834]} & .096 & .128 & .947 & .832 & .621 & .938 \\
& B3: Verb.\ Conf. & .733 {\tiny[.688,.778]} & .122 & .137 & .917 & .739 & .031 & .914 \\
& B4: Self-Consist. & .798 {\tiny[.759,.834]} & .096 & .128 & .947 & .832 & .621 & .938 \\
& B5: Embed Centr. & .807 {\tiny[.770,.846]} & .155 & .156 & .950 & .820 & .535 & .944 \\
& B6: LLM Agg. & .805 {\tiny[.765,.842]} & .102 & .110 & .953 & \best{.927} & .705 & \best{.956} \\
& M1: DUQ-LLM & .867 {\tiny[.831,.904]} & .062 & \best{.095} & \best{.965} & .886 & \best{.775} & .955 \\
& M2: DUQ-Embed & .798 {\tiny[.760,.837]} & .039 & .111 & .953 & .800 & .501 & .944 \\
& M3: DUQ-Learn & \best{.871} {\tiny[.833,.909]} & .074 & .097 & .958 & .869 & .781 & .948 \\
\midrule
\multirow{9}{*}{\rotatebox{90}{ARC-Chal.}}
& B1: Vote Count & .796 {\tiny[.704,.878]} & .089 & .056 & .976 & 1.00 & .982 & .981 \\
& B2: Vote Entropy & .794 {\tiny[.703,.876]} & .185 & .103 & .976 & 1.00 & .982 & .981 \\
& B3: Verb.\ Conf. & .760 {\tiny[.642,.860]} & .042 & .052 & .971 & 1.00 & .984 & .968 \\
& B4: Self-Consist. & .794 {\tiny[.703,.876]} & .185 & .103 & .976 & 1.00 & .982 & .981 \\
& B5: Embed Centr. & .778 {\tiny[.684,.867]} & .056 & .057 & .978 & 1.00 & .980 & .980 \\
& B6: LLM Agg. & \best{.868} {\tiny[.779,.947]} & .044 & \best{.046} & \best{.987} & 1.00 & \best{1.00} & \best{.990} \\
& M1: DUQ-LLM & .766 {\tiny[.635,.875]} & .024 & .048 & .974 & 1.00 & .982 & .972 \\
& M2: DUQ-Embed & .779 {\tiny[.678,.871]} & .039 & .050 & .975 & 1.00 & .984 & .973 \\
& M3: DUQ-Learn & .731 {\tiny[.602,.839]} & \best{.009} & .053 & .968 & 1.00 & .974 & .966 \\
\bottomrule
\end{tabular}
\caption{Complete results with bootstrap 95\% CIs for AUROC. \textbf{Bold}: best per benchmark-metric. DUQ = DiscoUQ.}
\label{tab:full-metrics}
\end{table*}

\section{Feature Importance Across Benchmarks}
\label{sec:feature-importance}

\begin{figure}[t]
\centering
\includegraphics[width=\columnwidth]{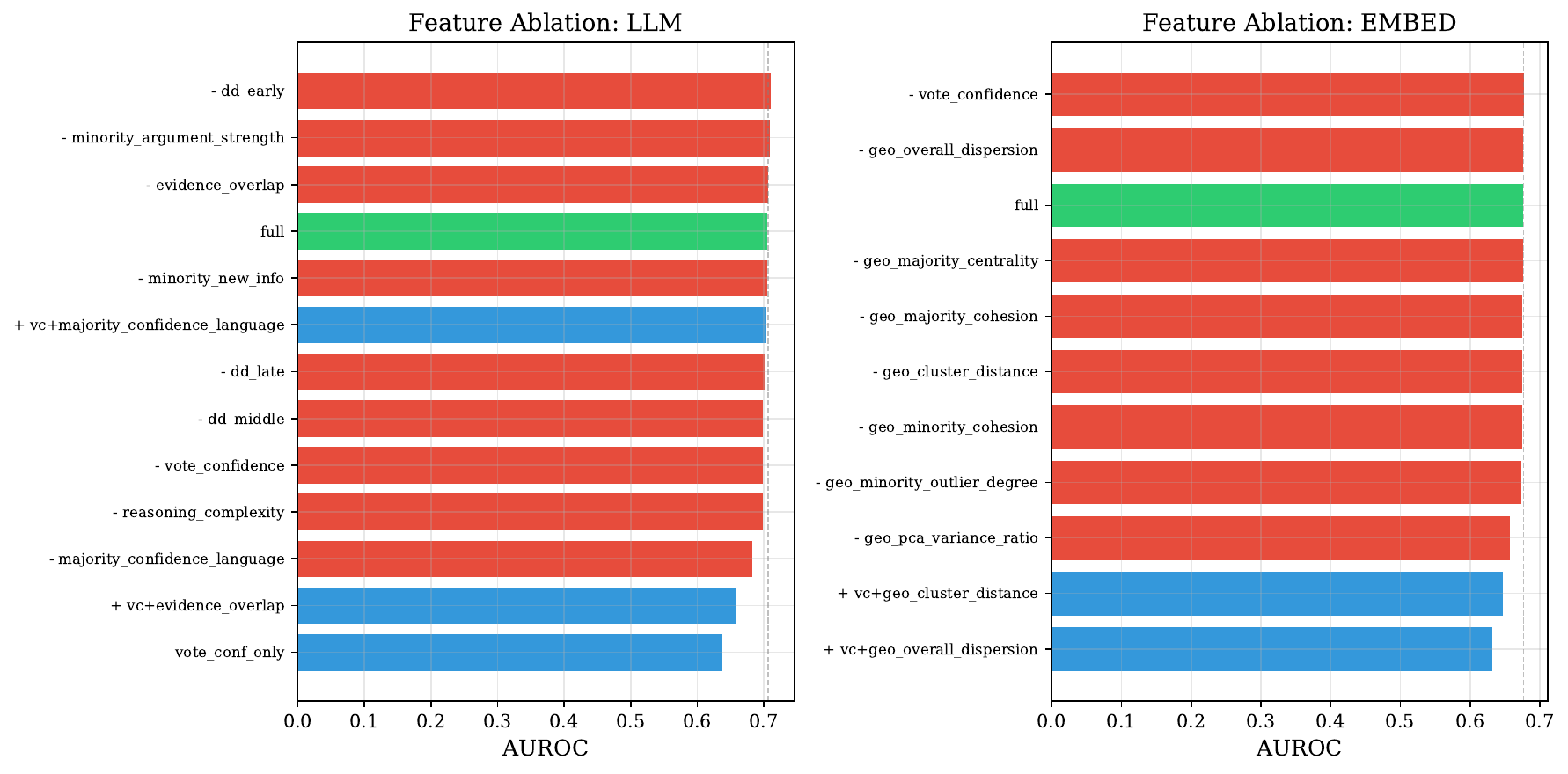}
\caption{Feature importance from ablation studies. Majority confidence language and reasoning complexity are the most important LLM structure features.}
\label{fig:feature-importance}
\end{figure}

Figure~\ref{fig:feature-importance} shows the ablation-based feature importance.

\section{Disagreement Profile Analysis}
\label{sec:disagree-profile}

We analyze the relationship between disagreement properties and accuracy in the weak tier by binning questions along two dimensions: evidence overlap (low/medium/high) and divergence depth (early/middle/late).

On StrategyQA, high evidence overlap combined with late divergence yields 60.3\% accuracy (116 questions), while low evidence overlap yields only 25.0\% accuracy (8 questions).
This pattern---shared evidence with late divergence as a positive signal---is consistent across benchmarks and confirms the informational value of structured disagreement features.

On MMLU's weak tier, we observe a clear gradient across M1 confidence groups: low-confidence questions have 17.9\% accuracy while high-confidence questions have 84.6\% accuracy, demonstrating that M1's structured features produce highly discriminative confidence scores even in the most challenging subset of questions.

\section{Experimental Details}
\label{sec:exp-details}

\paragraph{Agent prompts.}
Each of the 5 agents uses a distinct system prompt emphasizing a different reasoning style. For example, the Devil's Advocate agent is prompted: ``You are a critical thinker who always considers why the obvious answer might be wrong. Look for counterexamples, edge cases, and hidden assumptions.'' The full prompts are provided in our codebase.

\paragraph{Structure analysis prompt.}
The structure analysis uses a detailed prompt that presents all five agents' reasoning and asks for a JSON object with the six structure scores. Temperature is set to 0 for reproducibility. Analysis is only performed for non-unanimous votes (59\% of questions on average), reducing costs.

\paragraph{B6 LLM Aggregator.}
B6 concatenates all five agent responses (each truncated to 1,200 characters) and prompts the LLM to output a JSON with its answer and confidence. A system message (``You are a JSON-only responder'') ensures reliable parsing. Temperature is 0 and max tokens is 100.

\paragraph{Hyperparameters.}
For M3's MLP: 2 hidden layers of 32 units, ReLU activation, dropout 0.3, Adam optimizer with lr=$5{\times}10^{-4}$ and weight decay=$10^{-2}$, trained for up to 100 epochs with early stopping (patience 15), batch size 32.

\paragraph{Reproducibility.}
All random seeds are fixed at 42. Code and data will be made available upon publication.

\end{document}